\documentclass{article}

\usepackage{graphicx}
\usepackage{epstopdf}
\usepackage{indentfirst}
\usepackage{amsmath,amsthm,txfonts}
\usepackage{overpic}
\usepackage{float}
\usepackage{amsfonts}
\usepackage{cite}
\usepackage{dsfont}
\usepackage{bm}
\usepackage{multirow}

\usepackage[left=3cm,right=3cm,top=2.5cm,bottom=2.5cm]{geometry}
\usepackage{authblk}
\usepackage{multirow}
\usepackage{xcolor}
\usepackage{colortbl}
\usepackage{subfigure}
\usepackage{tikz}
\usetikzlibrary{calc}
\def\<{\left\langle}
\def\>{\right\rangle}

\twocolumn
\title{Binary output layer of feedforward neural networks for solving multi-class classification problems}
\author[a]{Sibo Yang}
\author[a]{Chao Zhang}
\author[a]{Wei Wu\thanks{Corresponding author: wuweiw@dlut.edu.cn}}
\affil[a]{\it \small School of Mathematical Sciences, Dalian
University of Technology, Dalian, 116024, China}

\begin{document}

\maketitle{}

\begin{abstract}

Considered in this short note is the design of output layer nodes of feedforward neural networks for solving multi-class classification problems with $r$ ($r\geq 3$) classes of samples. The common and conventional setting of output layer, called ``$one$-$to$-$one$ $approach$" in this paper,  is as follows: The output layer contains $r$ output nodes corresponding to the $r$ classes. And for an input sample of the $i$-th class ($1\leq i\leq r$), the ideal output is 1 for the $i$-th output node, and $0$ for all the other output nodes.   We propose in this paper a new  ``$binary$  $approach$": Suppose $2^{q-1}< r\leq 2^q$ with $q\geq 2$, then we let the output layer contain $q$ output nodes, and let the ideal outputs for the $r$ classes be designed in a binary manner.  Numerical experiments carried out in this paper show that our binary approach does equally good job as, but uses less output nodes than, the traditional one-to-one approach.
\end{abstract}

\bigskip

\textbf{Keywords:} \ Neural networks,\ Multi-class classification problems,\ One-to-one approach,\  Binary approach

\section{Introduction}
% no \IEEEPARstart
Learning efficiency and structural sparsification are two important issues in the study and application of neural networks.  The learning efficiency is mainly concerned with the choice of learning method so as to achieve good learning accuracy for the training samples and generalization (test) accuracy for the untrained samples [1-5]. The aim of structural sparsification is to use less numbers of nodes and connections (weights) without causing damage to the learning efficiency [6-10].

This short note considers the design of output layer nodes of feedforward neural networks for solving multi-class classification problem with $r$ ($r\geq 3$) classes of samples, and  proposes a novel approach using less output nodes than the conventional setting.

The common and conventional approach [11-18] for the design of output layer nodes is as follows: The output layer contains $r$ output nodes corresponding to the $r$ classes. For an input sample of the $i$-th class ($1\leq i\leq r$), the ideal output is 1 for the $i$-th output node, and $0$ for all the other output nodes. For example, for a classification problem with $r=4$, there are four output nodes in the output layer, and the ideal outputs for the four classes are $(1,0,0,0)$, $(0,1,0,0)$, $(0,0,1,0)$ and $(0,0,0,1)$, respectively. This approach  is called $one$-$to$-$one$ $approach$ in this paper.

We propose in this paper a new approach called $binary$  $approach$: Let $2^{q-1}< r\leq 2^q$ with $q\geq 2$. Then, we let the output layer contain $q$ output nodes, and let the ideal outputs for the $r$ classes be designed in a binary manner. For example, when $r=4$ and $q=2$, the output layer contains two output nodes, and the ideal outputs for the four classes are $(0,0)$, $(0,1)$, $(1,0)$ and $(1,1)$, respectively.

Numerical experiments carried out in this paper show that our binary approach does equally good job as, but uses less output nodes than,  the traditional one-to-one approach.

This paper is arranged as follows. In the next section, we describe the structure of the feedforward neural networks and the above mentioned two approaches of output layer setting. Then, in Section 3, we explain our ideas in terms of a few simple and  intuitive examples. Numerical simulations with five real world data sets are carried out in Section 4. Some conclusions are drawn in Section 5.

\section{Output layer setting for feedforward neural networks} % (fold)
\label{sec:}

% section  (end)
\subsection{Feedforward neural networks} % (fold)
\label{sub:}

% subsection  (end)
% subsection  (end)
% Ç°À¡Éñ¾­ÍøÂçµÄ½á¹¹Ò»°ã°üº¬ÊäÈë²ã¡¢Êä³ö²ã¡¢¼°Òþº¬²ã£¬Òþº¬²ã¿ÉÒÔÊÇÒ»²ã»ò¶à²ã¡£¸÷Éñ¾­ÔªÖ»½ÓÊÕÇ°Ò»²ãµÄÊä³ö×÷Ϊ×Ô¼ºµÄÊäÈ룬²¢ÇÒ½«ÆäÊä³ö¸øÏÂÒ»²ã¡£Ã¿Ò»¸öÉñ¾­Ôª¶¼¿ÉÒÔÓÐÈÎÒâ¶à¸öÊäÈ룬µ«Ö»ÔÊÐíÓÐÒ»¸öÊä³ö¡£
% Feedforward neural network generally includes the input, hidden and output layer.  Looking at the figure 1, each node receives one previous output as its input, then outputs it to the next layer. Each node can have a number of inputs, but only one output.

Let us begin with an introduction of a feedforward neural network with three layers. The node numbers of the input, hidden and output layers are $n$, $m$, and $p$ (see Figure 1), respectively. Let ${\{\mathrm x^h,\mathrm z^h\}}^{H}_{h=1}\subset R^n\times R^p$ be a given set of training samples, where $\mathrm x^h$ and $\mathrm z^h$ are the input and the corresponding ideal output of the $h$-th sample, respectively. Let $\mathrm V=(\mathrm V_{1}^T,\mathrm V_{2}^T,\cdots,\mathrm V_{m}^T)^T$ be the weight matrix connecting the input and the hidden layers, where $\mathrm V_{j}=(\mathrm V_{j1},\mathrm V_{j2},\cdots ,\mathrm V_{jn})$ for $j=1,2,\cdots ,m$. Let $\mathrm W=(\mathrm W_{1}^T,\mathrm W_{2}^T,\cdots,\mathrm W_{p}^T)^T$ be the weight matrix between the hidden and the output layers, where $\mathrm W_{k}=(\mathrm W_{k1},\mathrm W_{k2},\cdots ,\mathrm W_{km})$ for $k=1,2,\cdots ,p$.  $\{b_{1j}\}^{m}_{j=1}$ and $\{b_{2k}\}_{k=1}^p$ are the biases from input to hidden and from hidden to output layers, respectively. $f:R\to R$ denotes a given transfer function. In particular, we shall use the following  sigmoidal function in our numerical simulation:

\begin{equation}
  f(t)=\frac{1}{1+\mathrm {exp}(- t)}.
\end{equation}

\begin{figure}[H]
\centering{}

  \includegraphics[width=3in]{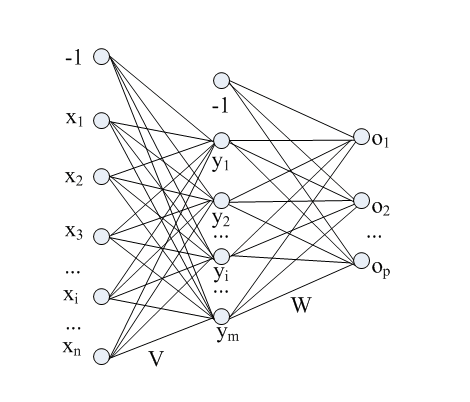}
  \caption {Structure of the feedforward neural networks}
      \label{fige1}
\end{figure}

For an input $x=(x_{1},\cdots,x_{n})^T\in R^n$, the output vector $y =(y_1,\cdots,y_m)^T$ of the hidden layer is given by
\begin{equation}
  y_{j}=f(\mathrm V_{j} \cdot \mathrm x -b_{1j})=f(\sum^{n}_{i=1} \mathrm V_{ji}  x_{i}-b_{1j}),\; j=1,\cdots,m,
\end{equation}
and the final output vector $o=(o_1,\cdots,o_p)^T\in R^p$ is given by
\begin{equation}
   o_{k}=f(\mathrm W_{k} \cdot \mathrm y -b_{2k})=f(\sum^{m}_{j=1} \mathrm W_{kj} y_{j}-b_{2k}),\;  k=1,\cdots,p.
\end{equation}

\subsection{Output layer settings} % (fold)
\label{sub:two_approaches_of_output_layer_setting}
% subsection two_approaches_of_output_layer_setting (end)
In the traditional setting of the output layer for solving a multi-class classification problem with $r$ classes of samples, if the input $x$ belongs to the $i$-th class of samples, the ideal output $z$ is

\begin{equation}
\begin{split}
z & =(z_{1},z_{2},\cdots,z_{i-1},z_{i},z_{i+1},\cdots ,z_{r})^T\\
&={(0,0,\cdots,0,1,0,\cdots,0)^T}.
\end{split}
\end{equation}
This setting of the output nodes  is called $one$-$to$-$one$ $approach$ in this paper.
An input $x\in R^n$ is classified into the $i$-th class if its network output (3) satisfies

\begin{equation}
\begin{split}
o & =(o_1,\cdots ,o_{i-1},o_{i},o_{i+1},\cdots ,o_p)^T\\
& \approx (0,\cdots ,0,1,0,\cdots ,0)^T.
\end{split}
\end{equation}
We say that the classification problem is successfully solved by the one-to-one approach if each input sample in the $i$-th class satisfies (5) for $i=1,2,\cdots,r$.
Due to our choice of the transfer function $f$, this implies that
\begin{equation}
  \left\{
   \begin{aligned}
   W_{k}\cdot y-b_{2k}>0,k=i,  \\
   W_{k}\cdot y-b_{2k}<0,k\neq i.\\
   \end{aligned}
   \right.
  \end{equation}
  Therefore, each sigmoidal function $f$ of an output node works like a hyperplane that separates one class of samples from all the other classes. An example is shown in Figure 2, where $m=2$, and the hyperplane becomes a line $l_i$,  such that the class $O_i$ and all the other classes $O_j$ $ (j\neq i)$ are divided by the line $l_i$.

  \begin{figure}[!h]
\centering{}

  \includegraphics[width=3in]{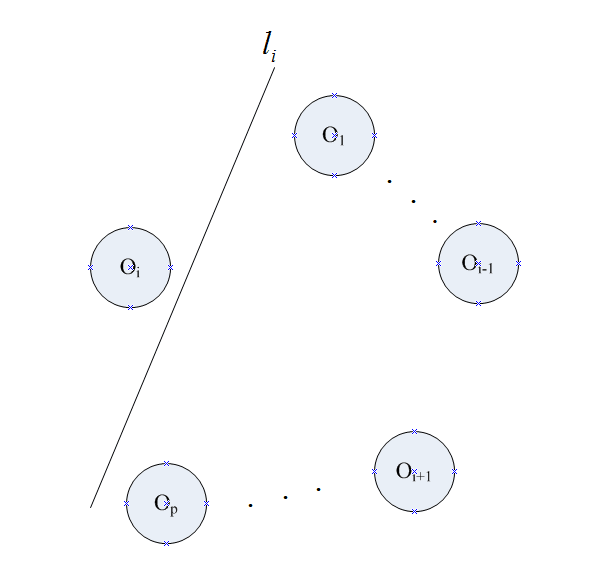}
  \caption {Class $O_i$ and all the other classes $O_{j}$ $(j\neq i)$ are divided by $l_i$.}
      \label{fige2}
\end{figure}

We propose in this paper another approach called $binary$  $approach$. Assume $2^{q-1}< r\leq 2^q$ with $q\geq 2$. Then,  $q$ output nodes are used in the output layer, and the ideal outputs for the $r$ classes are designed in a binary manner: The ideal output vector for the $i$-th class of samples is

 \begin{equation}
   z(i)=(z_{1},z_{2},\cdots,z_{q})^T,
\end{equation}
where
 \begin{equation}
   z_1z_2\cdots z_q=(i-1)_{2}.
\end{equation}
Here $(i-1)_{2}$ denotes the  binary number of $i-1$ with $z_j=0$ or $1$ for each $1\leq j\leq q$. Similarly, We say that the classification problem is successfully solved by the binary approach if each input sample in the $i$-th class satisfies the following condition for $i=1,2,\cdots,r$:
\begin{equation}
   (o_1, ,o_2,\cdots ,o_q)^T\approx z(i).
\end{equation}

\subsection{Learning algorithm} % (fold)
\label{sub:back_propagation_algorithm}

% subsection back_propagation_algorithm (end)
Now, assume that a training data set ${\{\mathrm x^h,\mathrm z^h\}}^{H}_{h=1}$ is given as mentioned in Subsection 2.1, and that ${\{\mathrm o^h\}}^H_{h=1}\subset R^p$ are the corresponding network outputs defined by (3). Define the error function as follows:

\begin{equation}
\begin{split}
& E(\mathrm W,\mathrm V)   =\frac{1}{2}\sum^H_{h=1} \| z_{h}-o_{h}\|^2  \\
&=\frac {1}{2}\sum^{H}_{h=1}\sum^{p}_{k=1}[z^{h}_{k}-f(\sum^{m}_{j=1}\mathrm W_{kj}f(\sum^{n}_{i=1}\mathrm V_{ji}x^{h}_{i}-b_{1j})-b_{2k})]^2.
\end{split}
\end{equation}

%\Delta \mathrm W_{kj}=-\eta $\eta>0$ is the learning rate,\Delta \mathrm V_{ji}\qquad
The aim of a learning algorithm is to choose the weight matrices  $\mathrm W$ and $\mathrm V$ so as to minimize the error function $E(\mathrm W, \mathrm V)$. To this end, we shall use the usual gradient descent algorithm.
Given arbitrary initial weight matrices $\mathrm W^{(0)}\in R^{p\times m}$ and $\mathrm V^{(0)}\in R^{m \times n}$, we update iteratively the weight matrices ${\mathrm W}^{(l)}=[\mathrm W^{(l)}_{kj}]$ and ${\mathrm V}^{(l)}=[\mathrm V^{(l)}_{ji}]$ as follows:

\begin{equation}
  \mathrm W^{(l+1)}_{kj}=\mathrm W^{(l)}_{kj}+\eta \frac{\partial E(W^{(l)},V^{(l)})}{\partial \mathrm W_{kj}},
\end{equation}

\begin{equation}
  \mathrm V^{(l+1)}_{ji}=\mathrm V^{(l)}_{ji}+\eta \frac{\partial E(W^{(l)},V^{(l)})}{\partial \mathrm V_{ji}}  ,
\end{equation}
where $l=0,1,2,...$; $k=1,2,...,p$; $j=1,2,...,m$; and $i=1,2,\cdots ,n$.

\section{Some simple and intuitive cases}

In this section, we try to explain our ideas by some intuitive observations in some simple cases.\\

{\bf Case 1.} First, let us consider  the simple case $r=2$. It is interesting that in this case everyone follows the  binary approach: Only a single output node is used,  and the two classes are labeled  by  the output values 1 and 0, respectively.  No one uses the one-to-one approach in this case by using two output nodes and labeling the two classes by outputs $(1,0)$ and $(0,1)$, respectively. Therefore, it seems that {\bf the binary approach, rather than the one-to-one approach, is a more natural extension for the output node setting from the simple case $r=2$ to the the general cases $r>2$}.

\bigskip

{\bf Case 2.} Let us consider a general case. Suppose the  one-to-one approach is successfully applied to  a classification problem with $r$-classes. Then, we can do equally well the same job after dropping out at least one output node. Let us take $r=4$ as an example: Originally, there should be four output nodes and the ideal outputs for the four classes are $(1,0,0,0)$, $(0,1,0,0)$, $(0,0,1,0)$ and $(0,0,0,1)$,  respectively. Then, we can simply drop out the last output node and set the new ideal output for the four classes be $(1,0,0)$, $(0,1,0)$, $(0,0,1)$ and $(0,0,0)$,  respectively.   This observation indicates that {\bf the one-to-one approach is not perfect in that one of its output node can be simply dropped out}.

\bigskip
{\bf Case 3.} Next, let us consider a four-class classification problem with $n$ dimensional input samples. Now, we assume the numbers of  the nodes for input, hidden and output layers are $n$, 2 and 4 respectively for the  one-to-one approach; and $n$, 2 and 2 for the binary approach. In the following  theorem, {\bf the binary approach is theoretically  proved to be at least as good as one-to-one approach in this special case.}\\

{\it{Theorem\ 3.1 Suppose that feedforward neural neural networks with two hidden nodes in the hidden layer are used for  solving a four-class classification problem. If the one-to-one approach can successfully solve the problem, then  the binary approach can also successfully solve the classification problem.}}\\

\begin{figure}[!h]
\centering{}
  \includegraphics[width=3in]{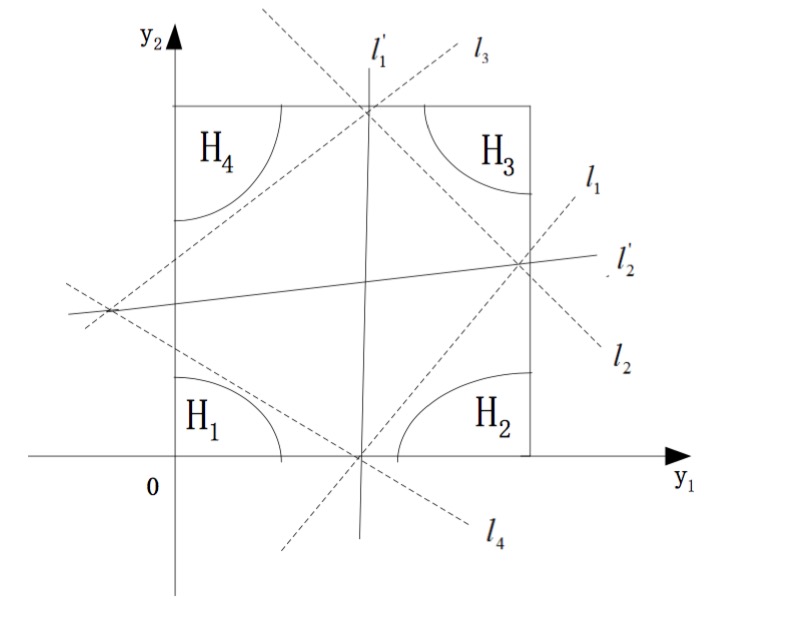}
\caption {Classification boundaries: solid lines stands for one-for-each approach; dotted lines for  binary approach.}
      \label{fige3}
\end{figure}

{\it Proof.} By the assumption of the theorem, there exists a set of weights such that the corresponding network of the one-to-one approach gives the desired outputs for the given data set. We note that this network maps the  input data set $P_I\subset R^n$ into a set $P_H\subset R^2$, and then maps the set $P_H$ into a set $P_O\subset R^4$. And  the four classes of input samples are mapped into four groups of points $P_{H1},\cdots,P_{H4}$ as illustrated in Figure 3, where $P_H=P_{H1} \bigcup P_{H2}\bigcup P_{H3}\bigcup P_{H4}$. (We remark that $P_H$ falls into the unit square of $R^2$ as shown in Figure 3 due to the choice of our transfer function $f(t)$ in (1).)   By recalling (4) and (6), we notice that the $i$-th ($1\leq i\leq 4$) output node acts like a line  $l_i$  that separates the point group $P_{Hi}$ from the other three point groups.

\bigskip

{Next, let us define two lines $l'_1$ and $l'_2$ as illustrated  in Figure 3. Obviously, these two lines divide the whole plane into four parts such that each part contains precisely a   $P_{Hi}$. As is well known, these two lines correspond to two output nodes which separate the four classes of samples from each other. This means that the corresponding binary approach can successfully classify the given data set as well. This completes the proof. $\square$}

\bigskip

{\bf Case 4.} Finally, let us give {\bf an example where the binary approach works better than the one-to-one approach}. Consider a four-class  classification problem with two dimensional input samples, of which the distribution is shown in Figure 4. Suppose we are using a two layer neural network (without the hidden layer) to solve this classification problem. In this case, we can easily use the two lines $l_1$  and $l_2$ shown in Figure 4 to separate the four classes, i.e., we can use a binary approach with only two layers (an input layer with $2$ nodes and an output layer with two nodes) to solve this classification problem. However, we can not do the similar thing by using the one-to-one approach with two layers (an input layer with $2$ nodes and an output layer with four nodes), since obviously there exists no line that can separate a class $I_{i}$ from the other three classes.\\

 \begin{figure}[H]
\centering{}

  \includegraphics[width=3in]{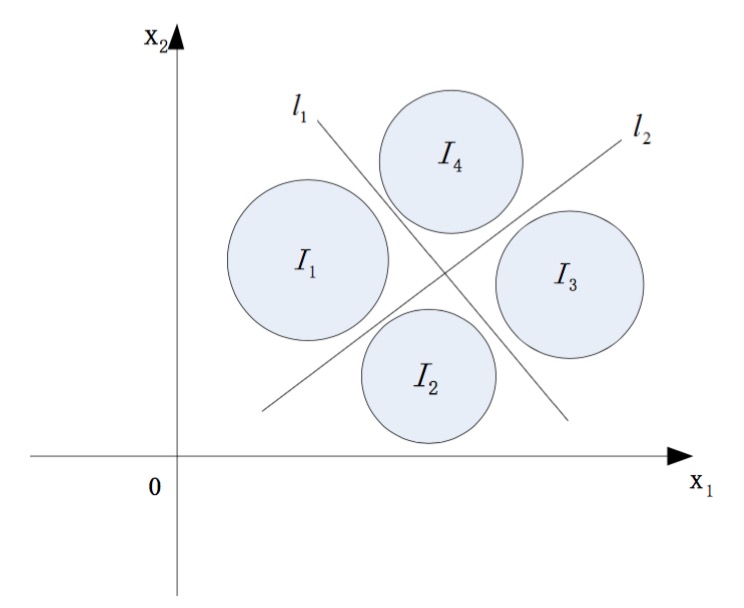}
  \caption {The binary approach can solve this classification problem, while the one-to-one approach can not.}
      \label{fige4}
\end{figure}{

\bigskip

{\bf Remark.} The above simple case studies explain and support our idea that the binary approach can work equally good as, or even better than, the  one-to-one  approach.  However, a theoretical and general  proof of the advantage of the binary approach over the  one-to-one  approach seems difficult or even impossible. In the next section, we shall turn to the numerical simulations to support  our idea.

\section{Numerical  examples}
% section three_numerical_experiments (end)
 In this section, we compare our binary approach with one-to-one approach on five real world classification problems. For each of the five data sets, the following five-fold cross validation technique [19-21] will be applied: The data set is divided randomly into five parts with equal (or nearly equal) number of samples. The network learning are carried out five times for the two approaches. At each time, one of the five parts are in turn chosen as the set of test samples, while the other four parts as the set of training samples. Then we re-start the process with re-arranged five parts of samples, and such process is repeated  twenty times. Altogether, for each approach-data pair, one hundred classification results are obtained.

\bigskip

 The ideal output value of an output node is either 1 or 0. When we evaluate the error between the ideal and real output values,  we shall use the following Fahlman's ``40-20-40" criterion [22]: The network output values between 0.00 and 0.40 of the output nodes are treated as 0, the values between 0.60 and 1.00 are treated as 1, and the values between 0.40 and 0.60 are treated as indeterminate and considered as incorrect.

\subsection{A four-class classification problem} % (fold)
First, we consider the four-class sensor drive diagnosis classification problem. This data set is publicly available from Machine Learning Respository at http://archive.isc.uci.edu. It comprises 21,276 input-output samples, each with 48 components. The ideal outputs of the four classes for the two approaches are shown in Table 1.

\begin{table}[!h]
\centering  % ±í¾ÓÖÐ
\begin{tabular}{lccc}  % {lccc} ±íʾ¸÷ÁÐÔªËضÔÆ뷽ʽ£¬left-l,right-r,center-c
\hline
class & one-to-one approach & binary approach \\ \hline  % \hline ÔÚ´ËÐÐÏÂÃæ»­Ò»ºáÏß
1 &(1,0,0,0) &(0,0)\\         % \\ ±íʾÖØпªÊ¼Ò»ÐÐ
2 &(0,1,0,0) &(0,1)\\        % & ±íʾÁеķָôÏß
3 &(0,0,1,0) &(1,0)\\
4 &(0,0,0,1) &(1,1)\\ \hline
\end{tabular}
\caption{Ideal outputs of the four classes.}
\end{table}

The network structures are 48-2-4 for the one-to-one approach and 48-2-2 for the binary approach. The learning rate $\eta$ is 0.06 and The maximum iteration number is 100.

The performances of the two approaches are shown in Table 2 and Figures 5-6. As we mentioned before, one hundred classification results are obtained for each of the two approaches. In the table and the figures, for instance, the ``average training accuracy" is over, and the ``highest training accuracy" is among, the one hundred training accuracies obtained.
From Table 2 it can be seen that the classification accuracies (average training accuracy, highest training accuracy, average test accuracy and highest test accuracy) of the binary approach are a little bit better than those of the  one-to-one approach. As shown in Figures 5-6, the values of the error function $E(W,V)$  for the binary approach are eventually lower than those for the one-to-one approach. Thus, in this example, the binary approach can do the job equally well as (actually a little bit better than), but use less output nodes than, the traditional one-to-one approach.

\begin{table}[!h]\footnotesize
\centering  % ±í¾ÓÖÐ
\begin{tabular}{|p{3cm}|p{1.5cm}|p{1.5cm}|} % {lccc} ±íʾ¸÷ÁÐÔªËضÔÆ뷽ʽ£¬left-l,right-r,center-c
\hline
\ & one-to-one approach & binary approach\\ \hline  % \hline ÔÚ´ËÐÐÏÂÃæ»­Ò»ºáÏß
average training accuracy &96.582\% &96.754\% \\
highest training accuracy &97.044\% &97.132\% \\
average test accuracy &96.336\% &96.548\% \\
highest test accuracy &96.871\% &96.934\% \\ \hline

\end{tabular}
\caption{Accuracies for the four-class classification problem.}
\end{table}

\begin{figure}[H]
\centering
\includegraphics[width=3in]{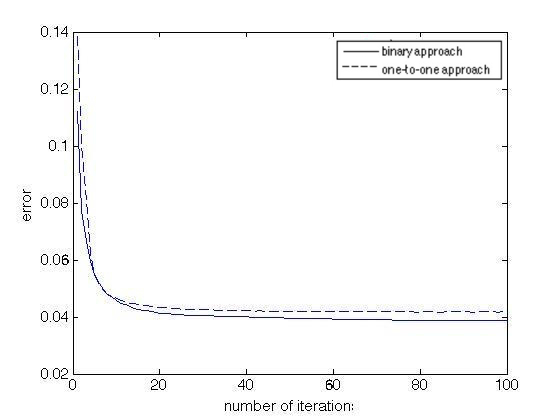}
\caption{Average values of the error function for the four-class classification problem.}
\label{fig:graph}
\end{figure}

\begin{figure}[!h]
\centering
\includegraphics[width=3in]{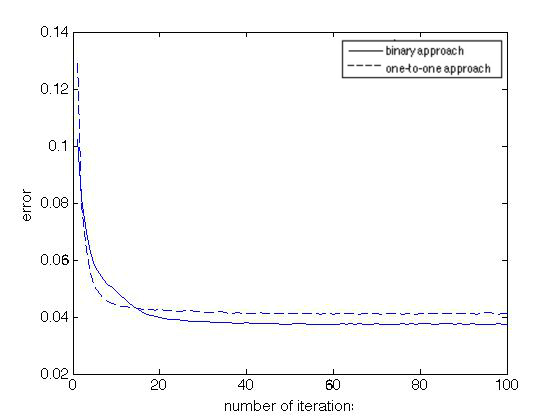}
\caption{Best values of the error function for the four-class classification problem.}
\label{fig:graph}
\end{figure}

\subsection{An eight-class classification problem} % (fold)
\label{sub:eight_classification_problem}

In this subsection, we consider the eight-class sensor drive diagnosis classification problem, which is also publicly available from Machine Learning Respository at http://archive.isc.uci.edu. The data set comprises 42,552 input-output samples, each with 48 components. The network structures are 48-3-8 for the one-to-one approach and 48-3-3 for the binary approach. For this classification problem, the learning rate $\eta$ is 0.1 and  the maximum iteration number is 500.

The performances of the two approaches are shown in Table 3 and Figures 7-8.
We observe that in this case, the classification Accuracies for the one-to-one approach are a little bit higher than those of the binary approach. Figures 5-6 show that the values of the error function   for the one-to-one approach are eventually lower  than those for the binary approach.

\begin{table}[!h]\footnotesize
\centering  % ±í¾ÓÖÐ
\begin{tabular}{|p{3cm}|p{1.5cm}|p{1.5cm}|}  % {lccc} ±íʾ¸÷ÁÐÔªËضÔÆ뷽ʽ£¬left-l,right-r,center-c
\hline
\ & one-to-one approach & binary approach\\ \hline  % \hline ÔÚ´ËÐÐÏÂÃæ»­Ò»ºáÏß
average training accuracy &92.255\% &90.530\% \\
highest training accuracy &92.844\% &91.032\% \\
average test accuracy &91.938\% &90.215\% \\
highest test accuracy &92.663\% &91.649\% \\ \hline

\end{tabular}
\caption{Accuracies for the eight-class classification problem.}
\end{table}

\begin{figure}[!h]
\centering
\includegraphics[width=3in]{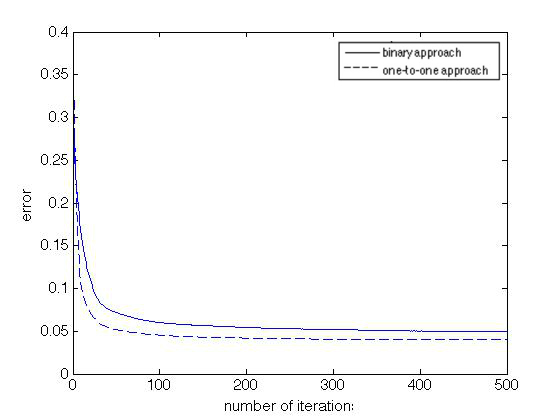}
\caption{Average values of the error function for the eight-class classification problem.}
\label{fig:graph}
\end{figure}

\begin{figure}[!h]
\centering
\includegraphics[width=3in]{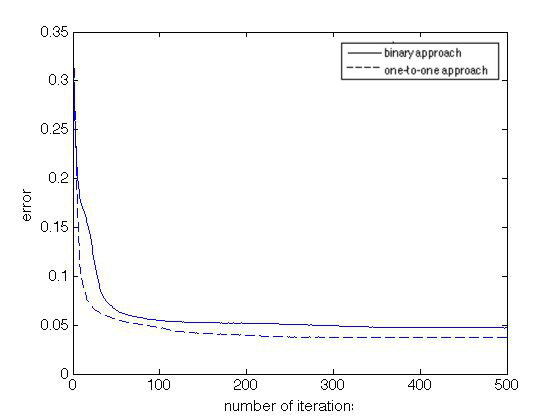}
\caption{Best values of the error function for the eight-class classification problem.}
\label{fig:graph}
\end{figure}

\subsection{A ten-class classificaion problem} % (fold)
\label{sub:ten_classificaion_problem}
Now, we consider the digit recognition problem, of which the aim is to classify the digit into ten categories (from 0 to 9). The data used here is publicly available on MNIST. The  data set comprises 70,000 input-output samples, each with 784 components.

The network structures are 784-4-10 for one-to-one approach and 784-4-4 for binary approach. For this classification problem, the learning rate $\eta$ is  0.08 and the maximum iteration number is 100.

The performances of the two approaches are shown in Table 4 and Figures 9-10.
In this example, the classification Accuracies for the binary approach are higher than those of the one-to-one approach. As shown in Figures 9-10, the values of the error function   for the binary approach are eventually lower than those for the one-to-one approach.

\begin{table}[!h]\footnotesize
\centering  % ±í¾ÓÖÐ
\begin{tabular}{|p{3cm}|p{1.5cm}|p{1.5cm}|} % {lccc} ±íʾ¸÷ÁÐÔªËضÔÆ뷽ʽ£¬left-l,right-r,center-c
\hline
\ & one-to-one approach & binary approach\\ \hline  % \hline ÔÚ´ËÐÐÏÂÃæ»­Ò»ºáÏß
average training accuracy &83.972\% &85.636\% \\
highest training accuracy &84.003\% &85.869\% \\
average test accuracy &83.536\% &85.202\% \\
highest test accuracy &83.691\% &85.307\% \\ \hline

\end{tabular}
\caption{Accuracies for the ten-class classification problem.}
\end{table}

\begin{figure}[!h]
\centering
\includegraphics[width=3in]{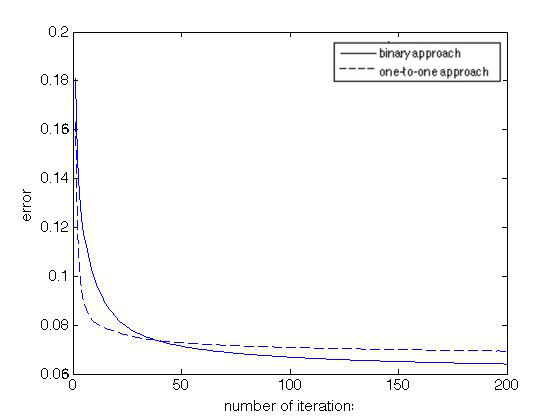}
\caption{Average values of the error function for the ten-class classification problem.}
\label{fig:graph}
\end{figure}

\begin{figure}[H]
\centering
\includegraphics[width=3in]{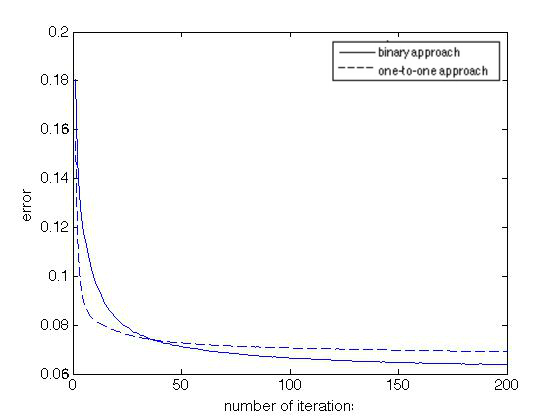}
\caption{Best values of the error function for the ten-class classification problem.}
\label{fig:graph}
\end{figure}

\subsection{An eleven-class classificaion problem}

In this subsection, we consider the eleven-class sensor drive diagnosis classification problem. The data set is publicly available from Machine Learning Respository at http://archive.isc.uci.edu. It comprises 58,509 input-output samples, each with 48 components.
The network structures are 48-4-11 for one-to-one approach and 48-4-4 for binary approach. For this classification problem, the learning rate $\eta$ is 0.1 and the maximum iteration number is 200.

The performances of the two approaches are shown in Table 5.
In this example, the classification accuracy of the binary approach is  higher than that of the one-to-one approach.

\begin{table}[!h]\footnotesize
\centering  % ±í¾ÓÖÐ
\begin{tabular}{|p{3cm}|p{1.5cm}|p{1.5cm}|}  % {lccc} ±íʾ¸÷ÁÐÔªËضÔÆ뷽ʽ£¬left-l,right-r,center-c
\hline
\ & one-to-one approach & binary approach\\ \hline  % \hline ÔÚ´ËÐÐÏÂÃæ»­Ò»ºáÏß
average training accuracy &82.672\% &83.835\% \\
highest training accuracy &83.359\% &84.626\% \\
average test accuracy &81.157\% &82.923\% \\
highest test accuracy &82.906\% &83.821\% \\ \hline

\end{tabular}
\caption{Accuracies for the eleven-class classification problem.}
\end{table}

\subsection{A twenty-six-class classificaion problem}

Now, we consider the letter recognition problem, of which the aim is to classify the letters into twenty-six categories (from A to Z). The data set is publicly available at http://www.ee.surrey.ac.uk/CVSSP/demos/-chars74k/. it comprises $1,016\times 26$ input-output samples, each with 784 components.
 The network structures are 784-5-26 for one-to-one approach and 784-5-5 for binary approach. For this classification problem, the learning rate $\eta$ is  0.08 and the maximum iteration number is 500.

The performances of the two approaches are shown in Table 6.
In this example, the classification Accuracies for the one-to-one approach are higher than those of the binary approach.

\begin{table}[!h]\footnotesize
\centering  % ±í¾ÓÖÐ
\begin{tabular}{|p{3cm}|p{1.5cm}|p{1.5cm}|}  % {lccc} ±íʾ¸÷ÁÐÔªËضÔÆ뷽ʽ£¬left-l,right-r,center-c
\hline
\ & one-to-one approach & binary approach\\ \hline  % \hline ÔÚ´ËÐÐÏÂÃæ»­Ò»ºáÏß
average training accuracy &82.686\% &80.653\% \\
highest training accuracy &83.284\% &81.376\% \\
average test accuracy &81.973\% &79.615\% \\
highest test accuracy &82.518\% &80.429\% \\ \hline

\end{tabular}
\caption{Accuracies for the twenty-six-class classification problem.}
\end{table}

\section{Conclusion} % (fold)
\label{sec:conclusion}
Considered in this short note is the design of output layer nodes of feedforward neural networks for solving multi-class classification problem with $r$ ($r\geq 3$) classes of samples. In this respect, the traditional $one$-$to$-$one$ approach uses $r$ output nodes such that for an input sample of the $i$-th class ($1\leq i\leq r$), the ideal output is 1 for the $i$-th output node, and $0$ for all the other output nodes. We propose
a novel approach called $binary$ approach: Let $2^{q-1}< r\leq 2^q$ with $q\geq 2$. Then  we let the output layer contain $q$ output nodes, and let the ideal outputs for the $r$ classes be designed in a binary manner. Therefore, less  output nodes are used in our binary approach.

Numerical simulations are carried out in this paper for solving five real world classification problems. Our binary approach performs better or slightly better  for three  classification problems, while the traditional one-to-one approach works better for the other  two classification problems. The differences of  the performances of the two approaches are not quite significant.  These numerical results show that, generally speaking,  our binary approach does equally good job as, but uses less output nodes than,  the traditional one-to-one approach.

 \section*{Acknowledgment}

 This work is partially supported by the National Natural Science Foundation of China: 61473059, 11401076 and 61473328; the Fundamental Research Funds for the Central Universities: DUT13-RC(3)068 and DUT17LK46; and Dalian High Level Talent Innovation Support Program: 2015R057.


\begin{thebibliography}{1}

\bibitem{IEEEhowto:kopka}
Safa N S, Ghani N A, Ismail M A. AN ARTIFICIAL NEURAL NETWORK CLASSIFICATION APPROACH FOR IMPROVING ACCURACY OF CUSTOMER IDENTIFICATION IN E-COMMERCE[J]. Malaysian Journal of Computer Science, 2014, 27(3):171-185.

\bibitem{IEEEhowto:kopka}
Kung J, Kim D, Mukhopadhyay S. On the Impact of Energy-Accuracy Tradeoff in a Digital Cellular Neural Network for Image Processing[J]. IEEE Transactions on Computer-Aided Design of Integrated Circuits and Systems, 2015, 34(7):1070-1081.

\bibitem{IEEEhowto:kopka}
Donald F. Specht and Philip D Shapiro. Generalization accuracy of probabilistic neural networks compared with backpropagation networks. Journal of Vector Ecology, 36(2): 426¨C36, 2011.

\bibitem{IEEEhowto:kopka}
Hassan M, Hamada M, Hassan M, et al. A Neural Networks Approach for Improving the Accuracy of Multi-Criteria Recommender Systems[J]. Applied Sciences, 2017, 7(9):868.

\bibitem{IEEEhowto:kopka}
Wang W C, Chau K W, Qiu L, et al. Improving forecasting accuracy of medium and long-term runoff using artificial neural network based on EEMD decomposition.[J]. Environmental Research, 2015, 139:46.

\bibitem{IEEEhowto:kopka}
Schmidhuber J. Deep learning in neural networks: an overview[J]. Neural Networks, 2015, 61:85-117.

\bibitem{IEEEhowto:kopka}
Tran J, Tran J, Tran J, et al. Learning both weights and connections for efficient neural networks[C]// International Conference on Neural Information Processing Systems. MIT Press, 2015:1135-1143.


\bibitem{IEEEhowto:kopka}
Wei Wu, Qinwei Fan, Jacek M. Zurada, Jian Wang, Dakun Yang, and Yan Liu. Batch gradient method with smoothing $L_{1/2}$ regularization for training of feedforward neural networks. Neural Networks, 50(2): 72¨C78, 2014.

\bibitem{IEEEhowto:kopka}
Qinwei Fan, Jacek M Zurada, and Wei Wu. Convergence of online gradient method for feedforward neural networks with smoothing $L_{1/2}$ regularization penalty. Neurocomputing, 131(9): 208¨C216, 2014.

\bibitem{IEEEhowto:kopka}
Song Han, Huizi Mao, and William J. Dally. Deep compression: Compressing deep neural networks with pruning, trained quantization and huffman coding. Fiber, 56(4): 3¨C7, 2016.

\bibitem{IEEEhowto:kopka}
Qingfeng Nie, Lizuo Jin, Shumin Fei, and Junyong Ma. Neural network for multi-class classification by boosting composite stumps. Neurocomputing, 149: 949¨C956, 2015.

\bibitem{IEEEhowto:kopka}
G. Sateesh Babu and S. Suresh. Meta-cognitive neural network for classification problems in a sequential learning framework. Neurocomputing, 81: 86¨C96, 2012.

\bibitem{IEEEhowto:kopka}
Simon S. Haykin. Neural networks and learning machines. China Machine Press,, 2009.

\bibitem{IEEEhowto:kopka}
Anas Quteishat, Chee Peng Lim, Jeffrey Tweedale, and Lakhmi C Jain. A neural network-based multi-agent classifier system. Neurocomputing, 72(79): 1639¨C1647, 2009.

\bibitem{IEEEhowto:kopka}
B. Gabrys and A. Bargiela. General fuzzy min-max neural network for clustering and classification. IEEE Transactions on Neural Networks, 11(3): 769, 2000.

\bibitem{IEEEhowto:kopka}
Eugene E. Clothiaux and Charles M. Bachmann. Neural Networks and Their Applications. Birkhuser, 2001.

\bibitem{IEEEhowto:kopka}
Guobin Ou and Lu Murphey Yi. Murphey, y.l.: Multi-class pattern classification using neural networks. pattern recogn. 40(1), 4-18. Pattern Recognition, 40(1): 4¨C18, 2007.

\bibitem{IEEEhowto:kopka}
Shigetoshi Shiotani, Toshio Fukuda, and Takanori Shibata. A neural network architecture for incremental learning. Neurocomputing, 9(2): 111¨C130, 1995.

\bibitem{IEEEhowto:kopka}
Barrow D K, Crone S F. Cross-validation aggregation for combining autoregressive neural network forecasts[J]. International Journal of Forecasting, 2016, 32(4):1120-1137.

\bibitem{IEEEhowto:kopka}
Jiang P, Chen J. Displacement prediction of landslide based on generalized regression neural networks with K -fold cross-validation[J]. Neurocomputing, 2016, 198:40-47.

\bibitem{IEEEhowto:kopka}
Nematzadeh Z, Ibrahim R, Selamat A. Comparative studies on breast cancer classifications with k-fold cross validations using machine learning techniques[C]// Control Conference. IEEE, 2015:1-6.


\bibitem{IEEEhowto:kopka}
N Ampazis and S. J Perantonis. Two highly efficient second-order algorithms for training feedforward networks. IEEE Transactions on Neural Networks, 13(5): 1064¨C1074, 2002.
\end{thebibliography}
\end{document}